\documentclass[11pt]{article}

\usepackage{acl}

\usepackage{times}
\usepackage{latexsym}

\usepackage[T1]{fontenc}

\usepackage[utf8]{inputenc}

\usepackage{microtype}

\usepackage{inconsolata}

\usepackage{graphicx}
\usepackage{amsmath}
\usepackage{amssymb}
\usepackage{booktabs}
\usepackage{multirow}
\usepackage[most]{tcolorbox}
\usepackage{enumitem}

\title{ActMem: Bridging the Gap Between Memory Retrieval and Reasoning in LLM Agents}

\author{
  Xiaohui Zhang$^1$, Zequn Sun$^1$, Chengyuan Yang$^1$, Yaqin Jin$^2$, Yazhong Zhang$^2$, Wei Hu$^{1,3}$ \\
  $^1$State Key Laboratory for Novel Software Technology, Nanjing University, China \\
  $^2$Alibaba Group, Hangzhou, China \\
  $^3$National Institute of Healthcare Data Science, Nanjing University, China \\
  \texttt{\{xhzhang, cyyang\}.nju@gmail.com, \{sunzq, whu\}@nju.edu.cn,} \\
  \texttt{\{jyq229592, zhangyazhong.zyz\}@taobao.com}
}

\newcommand{\categorycase}[4]{%
\begin{tcolorbox}[
  enhanced,
  breakable,
  colback=gray!8,
  colframe=gray!55,
  boxrule=0.6pt,
  arc=1mm,
  left=1.2mm,
  right=1.2mm,
  top=1mm,
  bottom=1mm,
  title={\textbf{#1}}
]
\small
\textbf{Question.} #2\par\smallskip
\textbf{History events.}
\begin{itemize}[leftmargin=*,itemsep=0pt,topsep=0pt,parsep=0pt]
#3
\end{itemize}
\textbf{Gold answer.} #4
\end{tcolorbox}
}

\begin{document}
\maketitle
\begin{abstract}
Memory management is essential for LLM agents in long-term interactions.
Current memory frameworks typically treat agents as passive ``recorders'' and retrieve information without understanding its deeper implications.
They may fail in scenarios requiring reasoning and complex decision-making.
To bridge this critical gap, we propose a novel actionable memory framework called ActMem that integrates memory retrieval with active causal reasoning.
ActMem transforms unstructured dialogue history into a structured causal and semantic graph.
By leveraging counterfactual reasoning and commonsense completion, it enables agents to deduce implicit constraints and resolve potential conflicts between past states and current intentions.
Furthermore, we introduce a comprehensive dataset ActMemEval to evaluate agent reasoning capabilities in logic-driven scenarios,
moving beyond the fact-retrieval focus of existing memory benchmarks.
Experiments demonstrate that ActMem significantly outperforms baselines in handling complex, memory-dependent tasks, paving the way for more consistent and reliable intelligent assistants.
\end{abstract}

\section{Introduction}

Large language models (LLMs) have shown remarkable capabilities in natural language understanding and generation, serving as the backbone for various agents.
To enable these agents to handle long-term interactions, effective memory management is essential~\cite{MemorySurvey}.
Conventionally, memory frameworks in LLM agents focus on expanding context windows or utilizing retrieval-augmented generation (RAG) to store, update and recall historical interactions.
They aim to improve \textit{the capacity of memory recall}, ensuring that the agent can retrieve specific information from a vast repository of past dialogues~\cite{memory_survey}.

However, a fundamental gap remains between simply remembering the past and effectively using it.
Existing memory management methods typically serve as a passive ``recorder,''
where the goal is to compress, summarize, and retrieve historical text~\cite{LightMem,ProMem}.
While RAG-based methods can fetch relevant dialogue snippets~\cite{rag_survey}, they often fail to understand the deeper implications of that memory for current decision-making.
In real-world applications, an agent acts as a human assistant, not just an ``archivist''~\cite{DailyAssistant}.
For instance, when a user proposes a new schedule, the agent should not only recall previous similar ones but also find event conflicts.
This requires the agent to go beyond memory retrieval and perform active reasoning based on historical states.
We define this capability as \textit{the utility of memory for action}.

This limitation is also reflected in current memory benchmarks.
Existing datasets, such as LoCoMo~\cite{lococmo}, LongMemEval~\cite{LongMemEval}, and HaluMem~\cite{HaluMem}, primarily evaluate the accuracy of memory retrieval through QA tasks.
They assess whether the agent can ``find'' the answer but overlook whether the agent can ``reason'' upon it.
They lack scenarios involving complex causal reasoning where the answer is not explicitly stated in the history but must be deduced.
Their evaluations fail to measure how well an agent can utilize memory to assist in current planning that may conflict with past state.

\begin{figure}[t]
\centering
\includegraphics[width=0.999\linewidth]{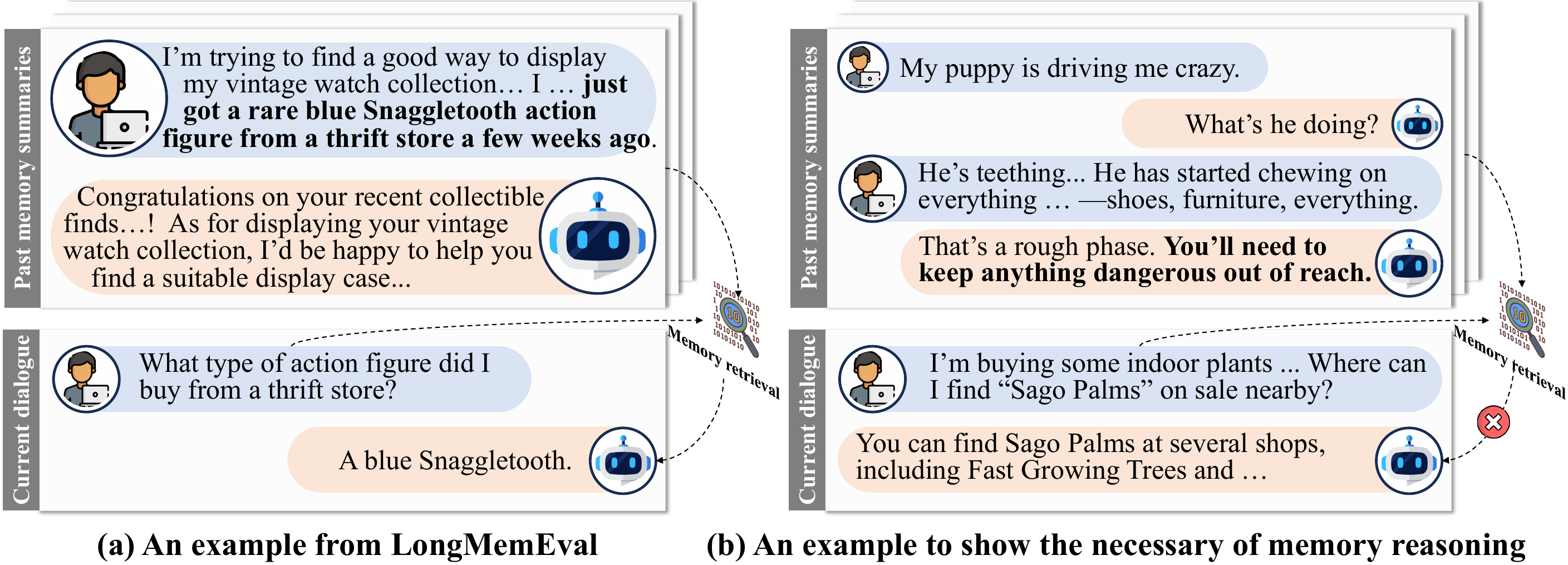}
\caption{A motivating comparison between memory retrieval and our memory reasoning.
(a) A memory retrieval example from LongMemEval.
(b) A memory reasoning example. 
The current query (buying Sago Palms) has little semantic overlap with the past memory (a teething puppy), but the agent should still infer the hidden conflict because Sago Palms are toxic to dogs.
}
\label{fig:example}
\end{figure}

To show the limitations of current benchmarks and the motivation of our work,
we present a comparison in Figure~\ref{fig:example}.
Subfigure (a) shows an example from LongMemEval.
The user's query explicitly matches a record in the past memory summaries.
The agent only needs to perform semantic or vectorized matching to retrieve the fact.
This is a task that evaluates the capacity of memory recall.

In contrast, Subfigure (b) shows a more realistic and challenging scenario that necessitates memory reasoning.
The user asks where to buy ``Sago Palms.''
A retrieval-only agent may simply return shopping information.
However, the dialogue history contains a critical information:
the user has a puppy that is ``teething'' and ``chewing on everything.''
Given that Sago Palms are highly toxic to dogs,
a responsible agent should infer the potential danger and intervene with a warning.
We find that LLMs generally possess this commonsense knowledge,
but current memory systems often fail to connect it with the stored history.
The inability to reason over memory can lead to conflicting or even dangerous advice.
This is the gap we aim to bridge. 
We want to turn the agent from a passive retriever into an active reasoner.

To bridge this gap, we propose ActMem, 
a memory framework that combines retrieval with causal reasoning.
It first extracts memory facts from raw dialogue history.
Then it clusters related facts and adds semantic edges between similar fact pairs.
After that, it mines causal relations of facts.
Furthermore, it performs counterfactual reasoning, commonsense completion and PMI-based filtering to refine cause edges.
These facts and edges form a memory knowledge graph (KG).
The KG helps the agent find associations, detect conflicts, and infer hidden constraints for current actions.
We also introduce a new dataset, ActMemEval, to evaluate this setting.
Unlike traditional retrieval benchmarks, it focuses on active reasoning and causal deduction.
In summary, the contributions of this paper are as follows:

\begin{itemize}
    \item We identify the critical limitation of current agent memory systems: the disconnect between memory retrieval and reasoning. We propose a shift in focus from the ``capacity of recall'' to the ``utility of memory for action.''

    \item We introduce ActMem, an actionable memory management framework. By integrating a memory KG with causal and semantic edges, ActMem enables agents to utilize past experiences for robust decision-making.

    \item We construct a comprehensive dataset ActMemEval to evaluate the reasoning capabilities of agents in long-term interactions, covering complex tasks that require logic-driven memory utilization.

    \item Extensive experiments demonstrate that ActMem significantly outperforms existing baselines in handling complex, logic-driven tasks. It also achieves competitive and even SOTA performance on conventional datasets.
\end{itemize}

We hope this work will pave the way for future research into more consistent and human-aligned intelligent assistants.
Our source code and dataset is available at GitHub.~\footnote{\url{https://github.com/nju-websoft/ActMem}}

\section{Related Work}

\subsection{Agent Memory Management}
Memory management is a core capability for intelligent agents in long-term interactions~\cite{MemorySurvey,memory_survey}.
Existing studies mainly fall into three groups: retrieval-centric memory, structured memory, and lightweight memory systems.

\smallskip\noindent\textbf{Retrieval-centric memory management.}
These methods mainly improve access to relevant history in long contexts.
Methods such as Mem0~\cite{Mem0} and MemoryBank~\cite{MemoryBank} follow a RAG-style paradigm~\cite{rag_survey}. 
They focus on storing, retrieving, and updating memories.
Their main goal is efficient and accurate recall from large memory banks.

\smallskip\noindent\textbf{Structured memory management.}
To overcome the limits of flat text, recent work explores structured memory formats.
Examples include tree-based methods like MemTree~\cite{MemTree} and graph-based methods like A-Mem~\cite{A-MEM}.
Other studies use hierarchical or compositional designs~\cite{Zep,G-Memory} to model dependencies across turns.
These structures often improve retrieval precision over plain text chunks.

\smallskip\noindent\textbf{Lightweight memory systems.}
Another line of work focuses on memory efficiency under limited token budgets.
LightMem~\cite{LightMem} uses hierarchical compression to summarize dialogue history and reduce storage and retrieval cost.
SimpleMem~\cite{simplemem2026} further adds semantic lossless compression, structured indexing, and adaptive retrieval to improve token usage.
These methods emphasize compact memory and efficient access. They are suitable for long-term interaction under limited computation.

\smallskip\noindent\textbf{Discussions.}
Despite this progress, a key gap remains. Recent surveys~\cite{memory_survey} show that most methods still treat memory as passive storage with fixed representations.
Whether they use RAG or trajectory replay, retrieval still depends heavily on surface semantic similarity. This makes it hard to detect hidden conflicts or adapt past experience to changing situations.
As a result, these methods may suffer from negative transfer or hallucination.
In contrast, ActMem uses an event-centric design that explicitly models logical structure and causal dependency. It turns memory from passive storage into an active reasoning engine.

\begin{figure*}[t]
\centering
\includegraphics[width=0.9\textwidth]{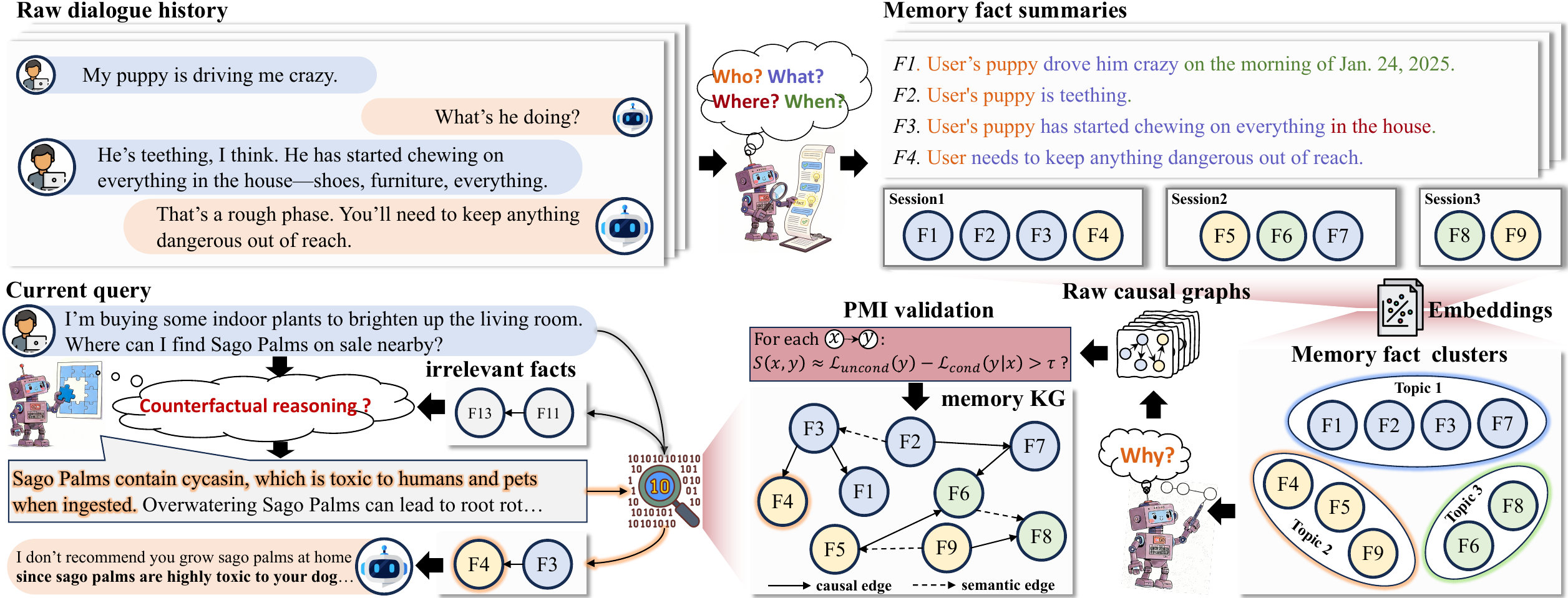}
\caption{The overview of our proposed framework.}
\label{fig:framework}
\end{figure*}

\subsection{Synthetic Evaluation Benchmark}
Synthetic benchmarks are widely used to evaluate LLMs in controlled settings.
They have moved from simple information retention to more complex reasoning.
Recent datasets like RULER~\cite{RULER} and BABILong~\cite{BABILong} test variable tracking and logical chaining. Datasets like LoCoMo~\cite{lococmo} and LongMemEval~\cite{LongMemEval} focus on long-term dialogue consistency and memory retrieval.
However, these datasets are mostly QA-oriented. They test passive retrieval more than active use of memory for decision-making.
We address this issue with ActMemEval. It evaluates the utility of memory for action via implicit constraints and causal conflicts.
We use human verification to improve reliability.

\section{The ActMem Framework}
\label{sec:method}

We present ActMem to bridge the gap between passive retrieval and active reasoning in LLM agents.
As illustrated in Figure~\ref{fig:framework}, the framework consists of four modules:
(1) memory fact extraction,
(2) fact clustering,
(3) memory KG construction,
and
(4) counterfactual-based retrieval and reasoning.
Formally, we define the raw dialogue history as $\mathcal{D} = \{u_1, a_1, u_2, a_2, \dots, u_t, a_t, \dots, u_N, a_N\}$, where $u_i$ and $a_i$ denote the user and agent utterances at turn $i$, respectively. Our goal is to transform $\mathcal{D}$ into a structured memory graph $\mathcal{G}$ to support complex decision-making.

\subsection{Memory Fact Extraction}
Raw dialogue history often contains redundancy and noise that hinder effective reasoning.
To address this, we first compress the raw history $\mathcal{D}$ into a set of atomic memory facts $\mathcal{F}$.
We use an LLM-based information extractor to parse the dialogue and generate declarative sentences containing key entities and facts following~\cite{LightMem}.
During extraction, we resolve pronouns in person or place mentions to explicit entities, and normalize relative time expressions into absolute timestamps based on the dialogue time.
We define an extraction function $f_{\text{ext}}(\cdot)$ powered by an LLM to map raw dialogues into a set of atomic facts $\mathcal{F}$.
For each dialogue turn $(u_t, a_t)$, we extract a subset of facts:
\begin{equation}
\resizebox{.88\columnwidth}{!}{$
\begin{split}
    \mathcal{F}_t = \text{LLM}(p_{\text{ext}}, u_t, a_t) = \{f_{t,1}, f_{t,2}, \dots, f_{t,k}\},
\end{split}
$}
\end{equation}
where $p_{\text{ext}}$ denotes the extraction prompt designed to identify entities and events. The global fact set is the union of all turns: $\mathcal{F} = \bigcup_{t=1}^N \mathcal{F}_t$.
Unlike traditional summarization which compresses the entire context into a single paragraph, our method focuses on \textit{atomicity}.
This granular extraction serves as the foundation for subsequent causal mining.

\subsection{Fact Clustering}
To model long-term interactions, we need to find relations between facts from different sessions.
Besides, mining causal edges by computing relations across all pairs of facts in the global history is time-consuming.
To mitigate this, we introduce a fact clustering module.
Let $\phi(\cdot)$ be an embedding function, e.g., Qwen3-Embedding-8B~\cite{Qwen3Embedding},
that maps a textual fact $f_i$ to a $d$-dimensional vector $\mathbf{h}_i \in \mathbb{R}^d$.
We compute semantic similarity with cosine similarity:
\begin{equation}
    \mathbf{h}_i = \phi(f_i), \ \ \ \text{sim}(f_i, f_j) = \frac{\mathbf{h}_i^\top \mathbf{h}_j}{\|\mathbf{h}_i\| \|\mathbf{h}_j\|}.
\end{equation}

To partition $\mathcal{F}$ into disjoint clusters $\mathcal{C} = \{C_1, \dots, C_K\}$ efficiently, we employ a single-pass incremental clustering algorithm.
We maintain a centroid vector $\mathbf{c}_k$ for each cluster. For each incoming fact $f_i$, we calculate its similarity with existing centroids. If the maximum similarity exceeds a clustering threshold $\delta$, the fact is assigned to the corresponding cluster; otherwise, a new cluster is initialized with $f_i$ as its centroid.
This strategy ensures that facts related to the same topic are consolidated dynamically.
It lets us mine causal relations inside local clusters, which reduces computation while keeping topical coherence.

\subsection{Memory KG Construction}
The core of ActMem is the construction of the memory KG, denoted as $\mathcal{G} = (\mathcal{V}, \mathcal{E})$, where nodes $\mathcal{V}$ correspond to memory facts and edges $\mathcal{E}$ represent causal or semantic relations.

\smallskip\noindent\textbf{Semantics edge generation.}
Following clustering, for each fact cluster, we calculate the pairwise similarity between facts and add a semantic edge between any two nodes whose similarity exceeds a semantic threshold $\tau$.

\smallskip\noindent\textbf{Raw causal edge generation.}
Within each cluster $C_k$ and each dialogue session,
we prompt the LLM to identify potential causal edges.
Let $\mathcal{H}_k$ be the set of candidate pairs $(f_i, f_j)$ identified by the model as potentially causal (i.e., $f_i \to f_j$):
\begin{equation}
    \mathcal{H}_k = \text{LLM}(p_{\text{cause}}, C_k).
\end{equation}

\smallskip\noindent\textbf{PMI-based hallucination filtering.}
A major challenge in using LLMs for causal mining is ``causal hallucination'' where the model identifies spurious correlations as causation.
To verify the candidate edges in $\mathcal{H}_k$, we employ a verification mechanism based on Pointwise Mutual Information (PMI).
It quantifies the information gain provided by the antecedent fact $f_i$ for generating the consequent fact $f_j$.
We define the verification score $S_{\text{PMI}}$ as the reduction in the negative log-likelihood of generating $f_j$ when conditioned on $f_i$:
\begin{equation}
\resizebox{.88\columnwidth}{!}{$
\begin{split}
    S_{\text{PMI}}(f_i, f_j) = \mathcal{L}_{\mathcal{M}}(f_j \mid T_{\text{uncond}}) - \mathcal{L}_{\mathcal{M}}(f_j \mid T_{\text{cond}}),
\end{split}
$}
\end{equation}
where $\mathcal{L}_{\mathcal{M}}(\cdot)$ denotes the cross-entropy loss of the target sequence $f_j$ computed by a pre-trained language model $\mathcal{M}$ (e.g., GPT2-Large).
To model the causal dependency, we design specific prompt templates for the input contexts:
(1) The \textit{conditional context} $T_{\text{cond}}$ constructs the sequence ``$f_i \text{. \textit{As a result}, } f_j$'';
(2) The \textit{unconditional context} $T_{\text{uncond}}$ uses a neutral prefix ``$\text{\textit{The fact is that} } f_j$''.
A positive $S_{\text{PMI}}$ means that $f_i$ clearly reduces the uncertainty of $f_j$.
We keep only pairs whose PMI score is above the threshold $\gamma$:
\begin{equation}
\resizebox{.88\columnwidth}{!}{$
\begin{split}
    \mathcal{E}_{\text{causal}} = \{(f_i, f_j) \in \bigcup \mathcal{H}_k \mid S_{\text{PMI}}(f_i, f_j) > \gamma\}.
\end{split}
$}
\end{equation}
Finally, we merge these validated causal edges with semantic edges to form the final memory graph $\mathcal{G}$.

\subsection{Counterfactual Retrieval and Reasoning}
Traditional RAG systems retrieve memories solely based on the semantic similarity to the user's current query $q$, often missing implicit constraints.
We propose a retrieval-reasoning-refinement loop to utilize the memory KG effectively.

\smallskip\noindent\textbf{Step 1: Initial retrieval.}
We retrieve an initial fact set $\mathcal{V}_{\text{init}} \subset \mathcal{V}$ by vector similarity to the query $q$:
\begin{equation}
    \mathcal{V}_{\text{init}} = \operatorname{TopK}(\operatorname{sim}(q, f_i), f_i \in \mathcal{V}).
\end{equation}

\smallskip\noindent\textbf{Step 2: Counterfactual reasoning.}
We then generate commonsense knowledge with a counterfactual question:
\begin{equation}
    k_{cs} = \text{LLM}(p_{\text{counter}}, q, \mathcal{V}_{\text{init}}).
\end{equation}
$p_{\text{counter}}$ can be briefly summarized as ``If the user does $q$, what negative consequences might occur considering $\mathcal{V}_{\text{init}}$ or general knowledge?''

\smallskip\noindent\textbf{Step 3: Refinement and response.}
We use $k_{cs}$ as an enhanced query for secondary retrieval over the memory graph.
First, we retrieve a seed set of nodes similar to $k_{cs}$:
\begin{equation}
    \mathcal{V}_{\text{seed}} = \operatorname{TopK}(\operatorname{sim}(k_{cs}, f_i), f_i \in \mathcal{V}),
\end{equation}
We then expand them with their structured graph information. Here, $\mathcal{N}_{\mathcal{G}}(v)$ denotes the neighbor nodes of $v$ in $\mathcal{G}$.
The refined set is
\begin{equation}
    \mathcal{V}_{\text{refine}} = \mathcal{V}_{\text{seed}} \cup \bigcup_{v \in \mathcal{V}_{\text{seed}}} \mathcal{N}_{\mathcal{G}}(v).
\end{equation}
The final context is $\mathcal{C}_{\text{final}} = \mathcal{V}_{\text{init}} \cup \{k_{cs}\} \cup \mathcal{V}_{\text{refine}}$.
The agent generates $ans$ given this context:
\begin{equation}
    ans = \text{LLM}(p_{\text{answer}}, q, \mathcal{C}_{\text{final}}).
\end{equation}
With this mechanism, ActMem turns the agent from a passive retriever to an active reasoner, ensuring the utility of memory for decision-making.

\section{The ActMemEval Benchmark}

In this section,
we introduce the pipeline to construct our ActMemEval and the evaluation results.

\begin{figure}[h]
\centering
\includegraphics[width=0.999\linewidth]{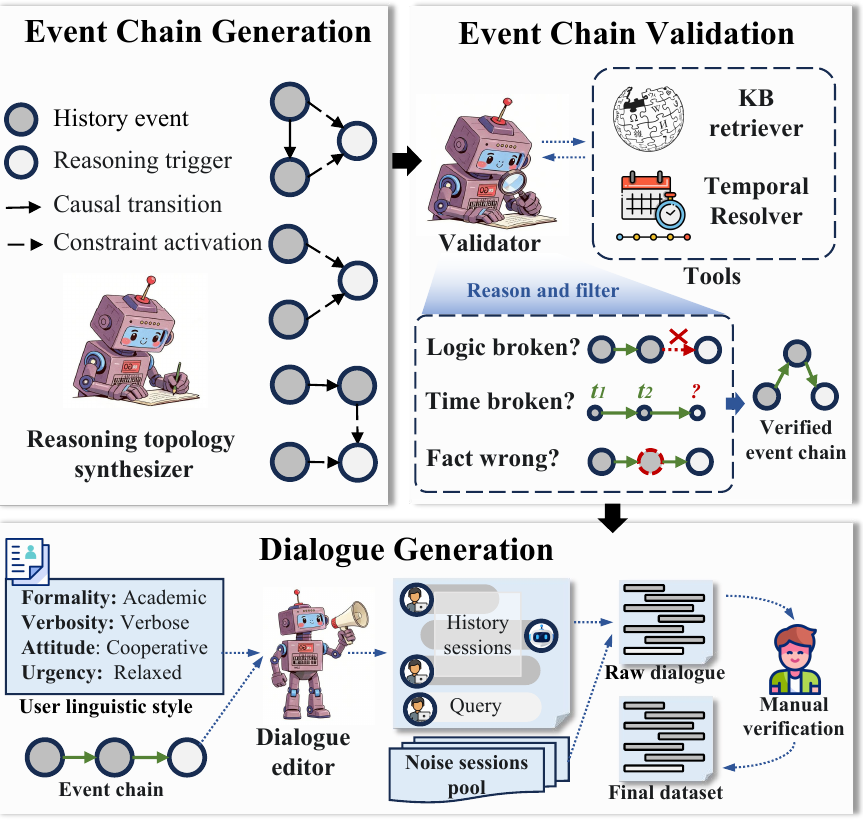}
\caption{Pipeline to construct our ActMemEval dataset.}
\label{fig:dataset}
\end{figure}

\subsection{Dataset Construction}
Figure~\ref{fig:dataset} shows the construction pipeline, which has three steps.
To avoid model self-bias and samples that are easy for the same model to solve, we use different models for data generation and validation. The generation model is also stronger than the model used in our method, which helps improve data quality. Specifically, we use Gemini 3.1 Pro for generation and Claude Sonnet 4.5 for validation.
To improve diversity, we synthesize six categories: safety-health risks, feasibility limits, time-space-procedure mismatches, access or availability gaps, preference incompatibility, and benefit-reuse opportunities. Detailed definitions are in Appendix~\ref{sec:app_dataset_categories}.

\smallskip\noindent\textbf{Step 1: Event Chain Generation.}
Unlike previous datasets that generate dialogues directly, we first synthesize the underlying event logic. We employ an advanced LLM (Gemini 3.1 Pro) as the event chain generator. It generates two types of nodes: history events and reasoning triggers.
For each sample, we randomly choose one category and generate the event chain accordingly.
History events describe the background facts, such as ``User adopted a puppy'' and ``Puppy is teething''.
Reasoning triggers represent the user's current intent, such as ``User wants to buy Sago Palms''.
The reasoning trigger links the history events and forms a complete reasoning chain.
In later dialogue generation, history events are converted into historical dialogues, while the reasoning trigger is converted into the current user query.

\smallskip\noindent\textbf{Step 2: Event Chain Validation.}
We then validate each chain from three aspects: logic, time, and knowledge. 
Claude Sonnet 4.5 checks logical consistency directly. 
For time and knowledge validation, we provide two external tools: \textit{Temporal Resolver} and \textit{KB retriever}. The validator calls them when needed. \textit{Temporal Resolver} normalizes time expressions into ISO dates and checks temporal order. \textit{KB retriever} queries Wikidata and Wikipedia and returns evidence snippets. The validator keeps only chains that are coherent, temporally valid, and knowledge grounded.

\smallskip\noindent\textbf{Step 3: Dialogue Generation.}
We convert the validated topologies into natural-language dialogues using a dialogue editor.
To reflect real-world diversity, we define a set of user linguistic styles along four dimensions: formality, verbosity, attitude, and urgency.
For each sample, we instantiate every node in the event chain as an anchor session.
The user in each anchor session follows one randomly sampled style combination.
Each anchor session reflects the facts in its corresponding node.
We then randomly sample noise sessions from UltraChat~\cite{ding-etal-2023-enhancing} and insert them between the anchor sessions to form a raw dialogue.
Finally, three NLP researchers manually verify each sample.
We keep only the samples with no disagreement.

\subsection{Dataset Evaluation}
\begin{figure}[t]
\centering
\includegraphics[width=0.9\linewidth]{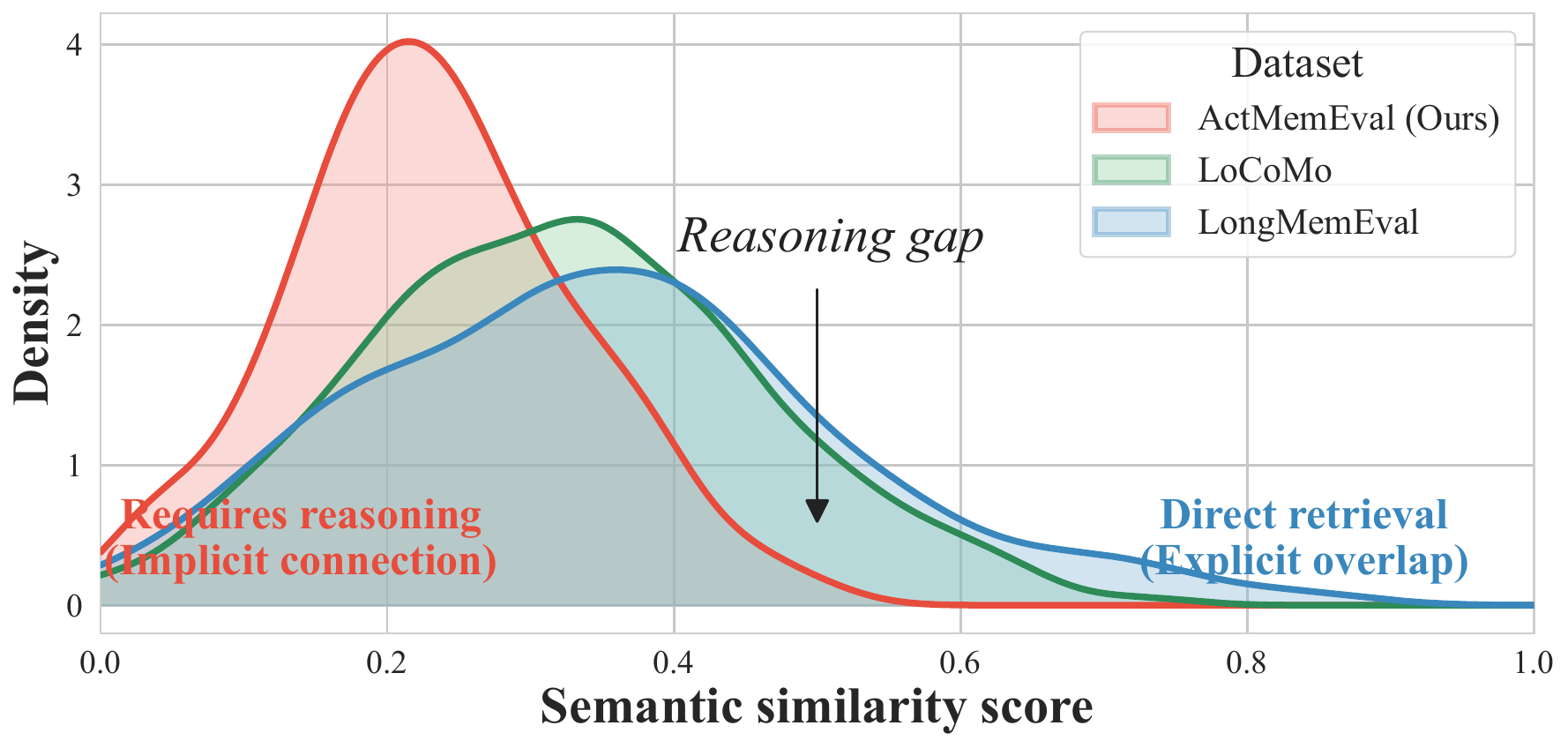}
\caption{Distribution of semantic similarity scores between each question and its supporting history context in ActMemEval, LoCoMo, and LongMemEval.}
\label{fig:sim}
\end{figure}

We evaluate the dataset from two aspects: context structure and query-history similarity.
We compare our ActMemEval with the retrieval-based benchmarks LongMemEval~\cite{LongMemEval} and LoCoMo.
Table~\ref{tab:dataset_stats} reports the basic statistics, and Figure~\ref{fig:sim} shows the similarity distribution between each question and its supporting history context.

\begin{table}[t]
\centering
\small
\setlength{\tabcolsep}{4pt}
\resizebox{\linewidth}{!}{
\begin{tabular}{lrrrrr}
\toprule
\textbf{Dataset} & \textbf{\#Q} & \textbf{Avg. Sess.} & \textbf{Avg. Tok.} & \textbf{Avg. Evid. Sess.} & \textbf{Avg. Evid. Pos.} \\
\midrule
ActMemEval & 327 & 67.25 & 120k & 2.48 & 34.53 \\
LongMemEval & 500 & 47.73 & 115k/1.5M & 1.90 & 25.16 \\
LoCoMo & 7412 & 27.20 & 10k & 1.29 & 14.14 \\
\bottomrule
\end{tabular}
}
\caption{Dataset statistics of ActMemEval and two related benchmarks. \#Q is the number of questions. Avg. Sess. is the average number of sessions. Avg. Tok. is the average number of tokens. Avg. Evid. Sess. is the average number of evidence sessions. Avg. Evid. Pos. is the average position of evidence sessions.}
\label{tab:dataset_stats}
\end{table}

Table~\ref{tab:dataset_stats} shows that ActMemEval is structurally denser. 
It has the largest average number of sessions and much longer contexts. 
Its evidence sessions are more numerous and appear later in the dialogue. 
Compared with LongMemEval and LoCoMo, ActMemEval spreads supporting clues across more sessions.

Figure~\ref{fig:sim} compares semantic similarity distributions. 
ActMemEval has the lowest average similarity (0.232) and LongMemEval is highest at 0.347. 
The ActMemEval curve is shifted left, 
which means the question is often semantically far from the key history. In other words, the evidence cannot be found by simple similarity retrieval.
Thus, ActMemEval is less retrieval-oriented and better tests memory reasoning.

\section{Experiments}

\subsection{Baselines}

We compare ActMem with six representative baselines:
NaiveRAG, MemoryBank~\cite{MemoryBank}, Mem0~\cite{Mem0}, A-Mem~\cite{A-MEM}, SimpleMem~\cite{simplemem2026}, and LightMem~\cite{LightMem}.
Detailed descriptions are provided in Appendix~\ref{sec:app_baselines}.

\subsection{Evaluation Metrics}
We employ two key metrics to assess both the \textit{precision of recall} and the \textit{effectiveness of reasoning}:

\begin{itemize}
\item \textbf{Retrieval accuracy:}
It tests whether the agent can recall the memory facts needed to answer the query.
Following~\cite{HaluMem}, retrieval accuracy is the recall rate of ground-truth evidence facts in our dataset.

\item \textbf{Question-answering accuracy:}
This metric measures the final utility of the memory.
It assesses whether the generated response correctly addresses the user intent while adhering to historical constraints.

\end{itemize}

\begin{table*}[t]
\centering
\small
\setlength{\tabcolsep}{3.2pt}
\resizebox{\textwidth}{!}{
\begin{tabular}{llccccccc}
\toprule
\textbf{LLM} & \textbf{Method} & \textbf{Overall} & \textbf{SHR} & \textbf{FL} & \textbf{TSP} & \textbf{AA} & \textbf{PI} & \textbf{BRO} \\
& & \textbf{Ret / QA} & \textbf{Ret / QA} & \textbf{Ret / QA} & \textbf{Ret / QA} & \textbf{Ret / QA} & \textbf{Ret / QA} & \textbf{Ret / QA} \\
\midrule
\multirow{7}{*}{\rotatebox{90}{\textbf{DeepSeek-V3}}}
& NativeRAG & \textbf{85.57} / 61.79 & \textbf{79.89} / 63.22 & \textbf{80.85} / \underline{55.32} & \textbf{92.86} / \underline{66.67} & \textbf{95.45} / \underline{68.18} & \textbf{91.67} / 53.33 & \textbf{86.11} / \underline{66.67} \\
& MemoryBank~\cite{MemoryBank} & 44.37 / 44.59 & 46.25 / 47.50 & 33.72 / 34.88 & 53.57 / 57.14 & 50.00 / 42.86 & 35.19 / 33.33 & 47.22 / 44.44 \\
& A-Mem~\cite{A-MEM} & 46.30 / 32.27 & 48.15 / 41.10 & 39.29 / 18.92 & 52.70 / 29.41 & 44.44 / 35.29 & 51.79 / 37.04 & 40.54 / 25.00 \\
& Mem0~\cite{Mem0} & \underline{78.07} / 41.80 & \underline{73.84} / 44.19 & \underline{78.72} / 34.04 & \underline{88.37} / 51.16 & \underline{75.00} / 45.45 & \underline{82.76} / 48.28 & 67.65 / 11.76 \\

& SimpleMem~\cite{simplemem2026} & 52.63 / 61.54 & 49.43 / 60.92 & 48.94 / 53.19 & 56.98 / 65.12 & 50.00 / 59.09 & 51.67 / 60.00 & 72.22 / \textbf{83.33} \\
& LightMem~\cite{LightMem} & 57.29 / \underline{62.35} & 60.34 / \underline{67.82} & 53.19 / 53.19 & 55.81 / 65.12 & 54.55 / 50.00 & 55.00 / \underline{63.33} & 63.89 / \underline{66.67} 
\\
\cmidrule(lr){2-9}
& ActMem & 71.66 / \textbf{76.52} & 71.51 / \textbf{82.56} & 71.28 / \textbf{72.34} & 75.58 / \textbf{86.05} & 68.18 / \textbf{72.73} & 66.67 / \textbf{66.67} & \underline{80.56} / 61.11 \\
\midrule
\multirow{7}{*}{\rotatebox{90}{\textbf{GPT-4o-mini}}}
& NativeRAG & \textbf{86.03} / 34.01 & \textbf{80.46} / 35.63 & \underline{79.79} / 19.15 & \textbf{94.19} / 39.53 & \textbf{93.18} / 45.45 & \textbf{96.67} / \underline{40.00} & \textbf{83.33} / 27.78 \\
& MemoryBank~\cite{MemoryBank} & 32.32 / 33.49 & 25.00 / 31.58 & 25.00 / \underline{31.71} & 33.82 / \underline{42.86} & 41.18 / 33.33 & 24.07 / 30.00 & 63.57 / \underline{33.33} \\
& A-Mem~\cite{A-MEM} & 45.29 / 18.42 & 38.06 / 22.73 & 34.72 / 13.89 & 48.61 / 13.89 & 45.83 / 33.33 & 58.33 / 12.50 & 71.88 / 18.75 \\
& Mem0~\cite{Mem0} & \underline{78.95} / 31.10 & \underline{79.58} / 38.03 & \textbf{81.71} / 29.75 & 80.77 / 30.77 & 72.50 / 25.00 & \underline{88.10} / 33.33 & 61.76 / 5.88 \\

& SimpleMem~\cite{simplemem2026} & 54.93 / 40.36 & 52.44 / \underline{48.78} & 46.34 / 26.83 & 60.53 / 42.11 & 57.50 / 45.00 & 51.79 / 32.14 & \underline{82.14} / \textbf{35.71} \\
& LightMem~\cite{LightMem} & 63.36 / \underline{40.49} & 63.79 / 44.83 & 58.51 / 27.66 & 67.44 / 41.86 & 63.64 / \textbf{54.55} & 63.33 / \underline{40.00} & 63.89 / 29.11 \\
\cmidrule(lr){2-9}
& ActMem & 76.92 / \textbf{66.80} & 75.29 / \textbf{79.31} & 77.66 / \textbf{65.96} & \underline{81.40} / \textbf{74.42} & \underline{77.27} / \underline{50.00} & 73.33 / \textbf{53.33} & 77.78 / \underline{33.33} \\
\bottomrule
\end{tabular}
}
\caption{Performance comparison on ActMemEval. We report retrieval accuracy (Ret) and QA accuracy (QA) for the overall dataset and for each category: safety-health risks (SHR), feasibility limits (FL), time-space-procedure mismatches (TSP), access or availability gaps (AA), preference incompatibility (PI), and benefit-reuse opportunities (BRO). Best results in each backbone block are highlighted in \textbf{bold}, and second-best results are \underline{underlined}.}
\label{tab:main}
\end{table*}

\subsection{Implementation Details}
\label{sec:implementation_details}

We employ two LLMs as backbones for reasoning and generation: DeepSeek-V3, representing high-performance open-source models, and GPT-4o-mini, representing efficient lightweight models.
For memory representation, we utilize Qwen3-Embedding-8B~\cite{Qwen3Embedding} to generate semantic vectors.
For hyperparameter settings, in the event clustering module, we set the clustering threshold $\delta$ to 0.2.
The semantic-edge threshold $\tau$ is set to 0.3.
For causal edge construction, the PMI validation threshold $\gamma$ is empirically set to 0.2.
During the retrieval phase, we set the number of retrieved memory facts to 20 for the initial retrieval step.
The number of counterfactual facts retrieved is set to 10.
In the refinement step, we dynamically retrieve additional facts based on the generated KG.
For the baseline configuration, we preserve each method's original final QA prompt and make minimal incremental modifications.
Specifically, we append brief instructions on reasoning to better adapt the baselines to our dataset.
This setting keeps the original prompting design intact while making the required reasoning process more explicit.
We use DeepSeek-V3 as the judge to evaluate retrieval results and generated responses, and we further verify its judgments through an agreement check with human annotation; see Appendix~\ref{sec:app_agreement}.

\subsection{Main Results}
Table~\ref{tab:main} presents the results on ActMemEval.
Although NaiveRAG achieves the highest retrieval accuracy by recalling extensive surface-level semantic matches,
its inferior QA performance indicates a failure to synthesize these facts with noise to deduce the implicit logical constraints required for QA.
Compared to LightMem,
ActMem achieves substantial improvements.
Specifically, when using DeepSeek-V3, it substantially improves QA accuracy over LightMem while also delivering consistently stronger retrieval quality across different categories.
A similar trend is observed with GPT-4o-mini.
In contrast, retrieval-based baselines exhibit relatively poor performance.
We attribute their failure to the reasoning gap.
These methods rely primarily on surface-level semantic matching, which struggles to retrieve historical events that are logically related but semantically distant from the current query.
For structured memory baselines, A-Mem underperforms ActMem because it uses the graph merely as a retrieval aid, while ActMem employs it as a reasoning framework with explicit causal and semantic edges.
In summary, the ability of ActMem to ``reason before retrieving'' allows it to identify implicit constraints that other methods miss, thereby achieving state-of-the-art performance in complex, action-oriented scenarios. A qualitative case study is provided in Appendix~\ref{sec:app_case_study}.

We also observe an interesting gap across different LLM backbones.
GPT-4o-mini attains high retrieval accuracy, yet its QA accuracy is markedly lower.
In contrast, DeepSeek-V3 delivers balanced and consistently superior results on both metrics.
We attribute this discrepancy to a reasoning bottleneck inherent in lightweight LLMs.
The lower QA accuracy suggests that GPT-4o-mini struggles to integrate and reason over these memory clues.
This phenomenon also reflects the difficulty of ActMemEval that requires reasoning.
With stronger reasoning ability, DeepSeek-V3 can use the retrieved context to resolve conflicts, while weaker LLMs may still fail even with correct evidence.

\begin{figure}[t]
\centering
\includegraphics[width=0.9\linewidth]{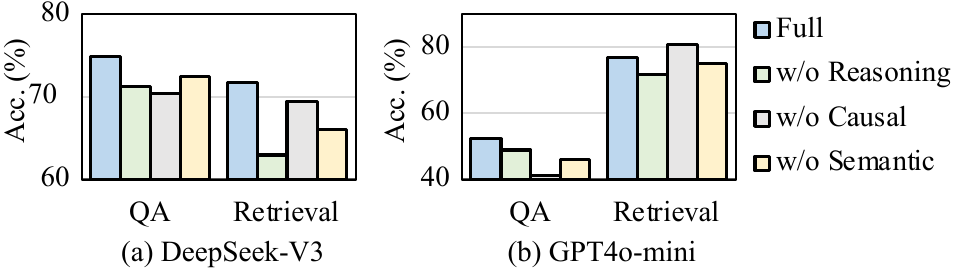}
\caption{Ablation study on counterfactual reasoning, causal and semantic edge construction.}
\label{fig:ablation}
\end{figure}

\subsection{Ablation Study}
To verify the contribution of each module,
we design three variants by removing key modules:
\begin{itemize}
\item \textbf{w/o Reasoning:} This variant removes the \textit{counterfactual reasoning} module. The agent retrieves memory facts directly from the memory KG without intermediate reasoning.
\item \textbf{w/o Causal:} It constructs the memory KG excluding \textit{causal edges}. 
The graph uses only semantic similarity to connect nodes.
\item \textbf{w/o Semantic:} This variant constructs the memory KG excluding \textit{semantic edges}. The graph consists only of extracted causal chains, making it a sparse causal network.
\end{itemize}

Figure~\ref{fig:ablation} presents the ablation results.
First, removing the reasoning module (\textit{w/o Reasoning}) with DeepSeek-V3 causes the largest drop in retrieval accuracy (about 9\%), confirming our core motivation.
In complex scenarios, relevant memories (e.g., ``owning a puppy'') may be semantically distant from the query (e.g., ``buying plants'').
Without counterfactual reasoning and commonsense completion, the agent fails to uncover such ``hidden'' evidence.
Second, removing causal edges (\textit{w/o Causal}) yields the lowest QA accuracy.
Even when some facts are retrieved, the absence of explicit causal links prevents the agent from forming coherent reasoning chains, leading to generic rather than conflict-aware decisions.
We observe that when using GPT-4o-mini, removing causal edges actually yields better retrieval performance than the full model.
We attribute this to the weaker reasoning capability of GPT-4o-mini, which may introduce errors during world-knowledge completion, adding more noise and degrading the retrieval quality of the full model.
Third, the \textit{w/o Semantic} variant also reduces retrieval accuracy, indicating that semantic edges serve as crucial connectors.
They ensure that topic-related but non-causal facts remain reachable, preserving the completeness of the retrieved context.
In summary, causal edges provide logical backbone, semantic edges ensure coverage, and the reasoning module guides inference.
Their combination gives ActMem strong performance. Additional analyses of PMI validation and semantic threshold sensitivity are in Appendices~\ref{sec:app_pmi} and~\ref{sec:app_threshold}, respectively.

\begin{figure}[t]
\centering
\begin{minipage}[t]{0.48\linewidth}
\centering
\includegraphics[width=\linewidth]{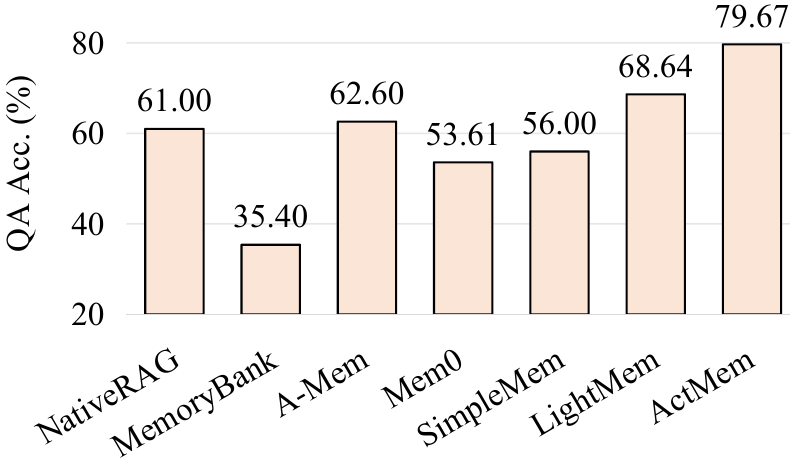}
\small (a) LongMemEval
\end{minipage}
\hfill
\begin{minipage}[t]{0.48\linewidth}
\centering
\includegraphics[width=\linewidth]{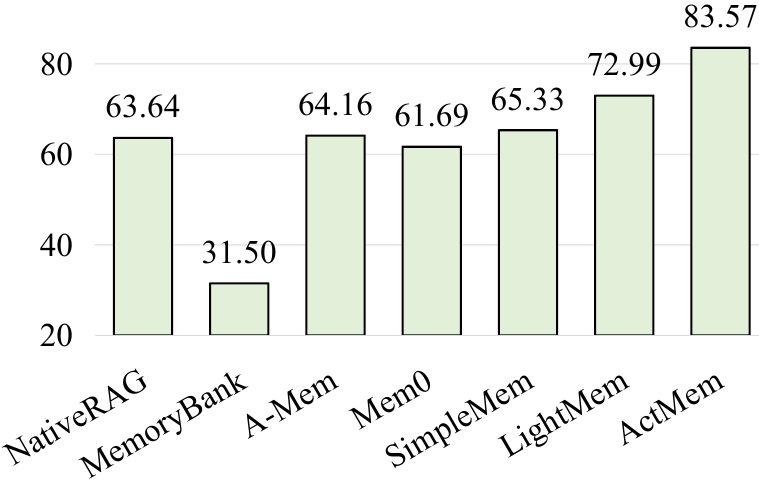}
\small (b) LoCoMo
\end{minipage}
\caption{Performance on conventional datasets.}
\label{fig:conventional_datasets}
\end{figure}

\subsection{Results on Conventional Datasets}

To test the generalization ability of ActMem, we run additional experiments on LongMemEval~\cite{LongMemEval} (LongMemEval-S) and LoCoMo~\cite{lococmo}.
For fairness, we use GPT-4o-mini as the backbone LLM for all methods and keep the evaluation setting consistent. ActMem uses the same parameters on all three datasets, without dataset-specific tuning.
As shown in Figure~\ref{fig:conventional_datasets}(a), ActMem achieves the best QA accuracy on LongMemEval.
It outperforms the strongest baseline LightMem, and also exceeds NativeRAG and A-Mem.
Figure~\ref{fig:conventional_datasets}(b) shows a similar trend on LoCoMo.
ActMem reaches 83.57\%, again ranking first among all methods.
These consistent gains on both datasets indicate that our structured memory KG is also effective in conventional retrieval-oriented settings.
By organizing scattered memories into connected fact structures, ActMem reduces retrieval noise and helps the model produce more accurate final answers.

\begin{figure}[t]
\centering
\includegraphics[width=0.9999\linewidth]{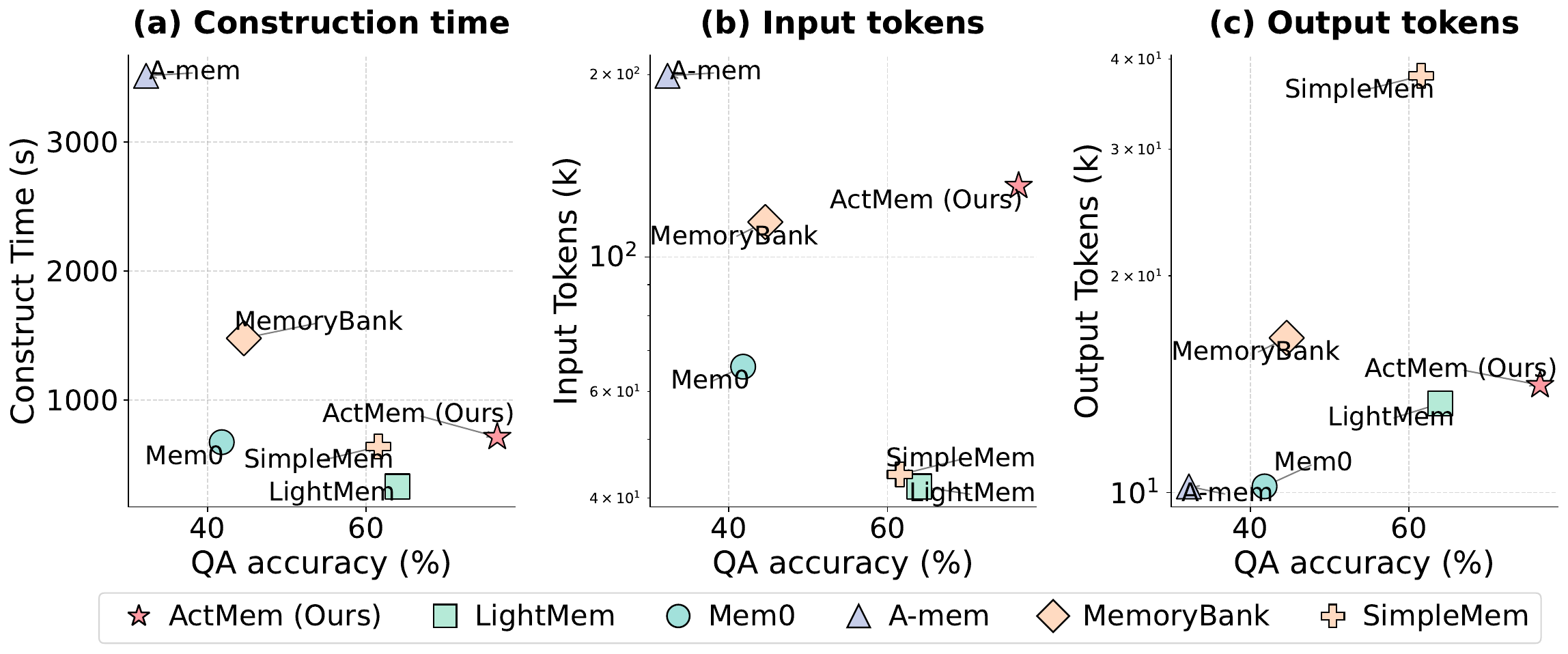}
\caption{Efficiency comparison using GPT-4o-mini.}
\label{fig:efficiency}
\end{figure}

\subsection{Efficiency Comparison}

Figure~\ref{fig:efficiency} compares efficiency and QA accuracy from multiple views.
Subfigure (a) shows that heavier frameworks such as A-Mem and MemoryBank have much higher construction cost, but still do not achieve competitive performance.
In contrast, ActMem achieves the best accuracy with moderate extra overhead.
Although it needs more construction time than lightweight methods such as LightMem and SimpleMem, it is still much more efficient than the most expensive baselines.
The second and third subfigures show the same trade-off in token usage.
ActMem uses more input tokens than compression-based methods because building the causal graph needs more context. However, this extra cost brings clear gains in answer quality.
At the same time, its output token usage stays close to LightMem and is much lower than SimpleMem. This suggests that the gain does not come from longer responses.
Overall, ActMem gives a good balance between reasoning performance and computational cost. 
It adds moderate overhead for causal and semantic reasoning, and converts that cost into strong performance.

\section{Conclusion and Future Work}

In this paper, we identify a key limitation of current agent memory systems: the gap between memory retrieval and reasoning.
We argue that high retrieval accuracy is not enough. Agents must not only recall past information, but also understand what it means for the present.
To address this issue, we propose ActMem, which transforms passive dialogue history into an active memory KG with causal and semantic edges.
We also introduce ActMemEval to evaluate the utility of memory for action.
Experiments on both our dataset and standard benchmarks show that ActMem outperforms baselines.

In future work, we plan to explore lightweight LLMs to improve KG construction efficiency without hurting reasoning quality.
We also plan to extend ActMem to multimodal scenarios and study fine-tuning strategies to improve causal reasoning.

\section*{Limitations}
Our method still has several limitations.
First, ActMem relies on LLM-based fact extraction, commonsense completion, and causal analysis, so errors in these stages may propagate to the final graph and response.
Second, constructing and refining the memory KG introduces additional time and token cost compared with simpler compression-based methods.
Third, ActMemEval focused on six predefined categories and cannot fully cover the diversity of real-world long-term interactions.

\bibliography{cite}

\appendix

\section{Baseline Details}
\label{sec:app_baselines}
We compare ActMem with six representative baselines:
\begin{itemize}
\item \textbf{NaiveRAG:} The standard baseline that retrieves the top-$k$ relevant dialogue snippets based on vector similarity and concatenates them into the context window.
\item \textbf{MemoryBank}~\cite{MemoryBank}: A summary-based memory framework that stores historical interactions as evolving summaries with a forgetting-aware update mechanism.
\item \textbf{Mem0}~\cite{Mem0}: A fact-level memory framework for personalized agents, focusing on continuous memory storage, updating, and retrieval.
\item \textbf{A-Mem}~\cite{A-MEM}: An agentic memory framework that models relations between memories and emphasizes association-driven retrieval.
\item \textbf{SimpleMem}~\cite{simplemem2026}: A lightweight lifelong memory framework based on semantic lossless compression, structured indexing, and adaptive retrieval.
\item \textbf{LightMem}~\cite{LightMem}: A lightweight memory framework that uses hierarchical compression and summary-based retrieval to reduce token cost.
\end{itemize}

\section{ActMem Case Study}
\label{sec:app_case_study}
Table~\ref{tab:examples} presents a case study from our dataset.
The user asks for a location to buy ``Sago Palms.''
Although the query appears simple, it contains a critical implicit constraint in the dialogue history: the user owns a puppy that is ``teething'' and ``chewing on everything.''
As Sago Palms are toxic to dogs,
an intelligent agent should detect this risk and issue a warning.
LightMem fails to retrieve the relevant evidence about the puppy, instead surfacing irrelevant noise such as ``high-end hotels'' and ``cooking meals.''
It treats the query as a straightforward information-seeking task and directly recommends where to buy the plants.
In contrast, ActMem successfully bridges this reasoning gap.
Through the counterfactual reasoning, the agent first infers the potential consequence of the user's intent (i.e., that Sago Palms are toxic).
This knowledge becomes a semantic anchor that guides retrieval of the specific memory that the user's puppy is chewing on everything.
By integrating this retrieved history with world knowledge,
ActMem identifies the conflict and produces a safety-aware response, advising against the purchase.
This case highlights our ability to transform memory from passive storage into an active reasoning engine for robust decision-making.

\begin{table*}[t]
\centering
\small
\resizebox{\textwidth}{!}{
\begin{tabular}{llllll}
\toprule
 & \textbf{Query} & \textbf{History} & \textbf{ActMem} & \textbf{Retrieved memory} & \textbf{Reasoning} \\
\midrule
\multirow{2}{*}{\rotatebox[origin=c]{90}{\parbox[c]{2.5cm}{\centering LightMem}}}
& \begin{minipage}[t]{2.2cm}
I'm buying some indoor plants to brighten up the living room. Where can I find ``Sago Palms'' on sale nearby?
\end{minipage}
& \begin{minipage}[t]{5cm}
[User]: My puppy is driving me crazy.

[Assistant]: What's he doing?

[User]: He's teething, I think. He has started chewing on everything in the house---shoes, furniture, everything.

[Assistant]: That's a rough phase. You'll need to keep anything dangerous out of reach.
\end{minipage}
& \begin{minipage}[t]{3.8cm}
I wouldn't recommend a Sago Palm... You mentioned last week that your new puppy is chewing on everything, and Sago Palms are extremely toxic---often fatal---to dogs if ingested. Can I suggest a pet-safe alternative ... ?
\end{minipage}
& \begin{minipage}[t]{3cm}
User is looking for high-end hotels in Narva that offer spa services or other luxury amenities.

User wants to make the most of their time in Narva and is seeking suggestions ...
\end{minipage}
& \begin{minipage}[t]{3.9cm}
... This is a great way to brighten up your living room and improve your environment! \textcolor{red}{Given that you've been ... planning a trip to Narva and cooking healthier meals ... Here are some steps you can take to find Sago Palms nearby:
...
}
\end{minipage} \\
\midrule
\multirow{2}{*}{\rotatebox[origin=c]{90}{\parbox[c]{2.5cm}{\centering ActMem}}}
& \begin{minipage}[t]{2.2cm}
I'm buying some indoor plants to brighten up the living room. Where can I find ``Sago Palms'' on sale nearby?
\end{minipage}
& \begin{minipage}[t]{5cm}
[User]: My puppy is driving me crazy.

[Assistant]: What's he doing?

[User]: He's teething, I think. He has started chewing on everything in the house---shoes, furniture, everything.

[Assistant]: That's a rough phase. You'll need to keep anything dangerous out of reach.
\end{minipage}
& \begin{minipage}[t]{3.8cm}
I wouldn't recommend a Sago Palm... You mentioned last week that your new puppy is chewing on everything, and Sago Palms are extremely toxic---often fatal---to dogs if ingested. Can I suggest a pet-safe alternative ... ?
\end{minipage}
& \begin{minipage}[t]{3cm}
User is interested in creative ways to use zucchini in cooking.

\textcolor{blue}{User's puppy has started chewing on everything in the house, including shoes and furniture.
}

...
\end{minipage}
& \begin{minipage}[t]{3.9cm}
\textcolor{blue}{
... The user's request to buy Sago Palms directly conflicts with their current situation:
The user has a puppy. World knowledge indicates that Sago Palms are highly toxic to dogs and cats ...
}
\end{minipage} \\
\bottomrule
\end{tabular}}
\caption{A case study from our ActMemEval benchmark.}
\label{tab:examples}
\end{table*}

\section{PMI Verification Case Study}
\label{sec:app_pmi}
To further illustrate the effect of PMI validation, we randomly sampled five positive examples and five negative examples from candidate causal pairs.
Tables~\ref{tab:pmi_positive_examples} and~\ref{tab:pmi_negative_examples} report their PMI scores.
The positive examples all exceed the threshold $\gamma = 0.2$ and correspond to clear causal relations, while the negative examples remain near zero or below zero and are filtered out.
This case study shows that PMI helps suppress weak or spurious causal pairs.

\begin{table*}[t]
\centering
\small
\setlength{\tabcolsep}{3pt}
\begin{tabular}{p{1cm}p{7.7cm}p{6.3cm}}
\toprule
\textbf{PMI} & \textbf{Cause} & \textbf{Effect} \\
\midrule
0.9163 & User had a scary reaction where their hands swelled up like balloons after cleaning the bathroom yesterday. & User's doctor advised them to banish all natural rubber from the house due to a reaction. \\
0.4278 & User's iron levels are critically low, and their supplements are not absorbing effectively. & User's doctor advised them to eat steak twice a week to avoid needing hospital infusions. \\
1.4725 & User's grandmother's mobility has worsened after a fall, and she is now permanently using a wheelchair. & User's family is adjusting travel arrangements for their grandmother due to her permanent use of a wheelchair. \\
1.3756 & The railway workers' union announced a total 24-hour strike for tomorrow. & No trains will be running tomorrow due to the railway workers' union strike. \\
1.2329 & User's professor emailed that all proctored exams must go to the North Campus facility during the downtown closure. & User needs to book a slot at the North Campus facility for May 18th at 9 AM. \\
\bottomrule
\end{tabular}
\caption{Randomly sampled positive examples retained by PMI validation.}
\label{tab:pmi_positive_examples}
\end{table*}

\begin{table*}[t]
\centering
\small
\setlength{\tabcolsep}{3pt}
\begin{tabular}{p{1.1cm}p{6cm}p{7.5cm}}
\toprule
\textbf{PMI} & \textbf{Cause} & \textbf{Effect} \\
\midrule
-0.0655 & User used 2 tablespoons of vegetable oil. & User thinks the vegetable stir-fry recipe sounds delicious and is seeking a protein option to pair with it, as well as tips for making the veggies really crispy. \\
0.0738 & The application will have a smooth, modern user interface. & User is seeking more information regarding the customizable intervals in the time-lapse photography application. \\
0.0859 & User found the comprehensive guide on how to make a room diffuser very helpful. & User wants warm and cozy essential oil combinations to help create a relaxing atmosphere in their home. \\
0.0646 & User's doctor said they have to eat a steak twice a week or they will need hospital infusions. & User felt like a traitor to the cause on March 22, 2023. \\
0.0923 & User is really excited to get involved in reforestation efforts. & User is going to start researching local organizations and events to make a difference in their community. \\
\bottomrule
\end{tabular}
\caption{Randomly sampled negative examples filtered out by PMI validation.}
\label{tab:pmi_negative_examples}
\end{table*}

\section{Analysis of Semantic Similarity Thresholds}
\label{sec:app_threshold}
We investigate the impact of the semantic similarity threshold on QA accuracy and graph density metrics.
As presented in Table~\ref{tab:semantic_threshold_analysis},
we observe a clear correlation between graph connectivity and QA accuracy.
A lower threshold (e.g., 0.3) yields the optimal QA accuracy of 76.52\%.
This improvement stems from the richer retrieved context: a relaxed threshold preserves more semantic edges, leading to a substantially higher average node degree (20.48) and more retrieved KG facts (5.21). The resulting dense connectivity provides the agent with broader semantically relevant background information, reducing the risk of missing critical clues.
In contrast, a stricter threshold (e.g., 0.7) produces a much sparser graph, with the average degree dropping to 2.34. Such aggressive pruning removes potential semantic associations, resulting in insufficient context retrieval (only 0.29 facts on average) and a corresponding decline in QA accuracy. These results highlight the trade-off between graph sparsity and information recall.

\begin{table}[t]
\centering
\resizebox{\linewidth}{!}{
\begin{tabular}{cccc}
\toprule
\textbf{Threshold} & \textbf{QA acc.} & \textbf{Avg. node degree} & \textbf{Avg. retrieved KG facts} \\
\midrule
0.3 & 76.52 & 20.48 & 5.21 \\
0.5 & 74.89 & 5.07 & 1.24 \\
0.7 & 75.30 & 2.34 & 0.29 \\
\bottomrule
\end{tabular}}
\caption{QA accuracy and graph density metrics with different semantic similarity thresholds.}
\label{tab:semantic_threshold_analysis}
\end{table}

\section{Dataset Categories}
\label{sec:app_dataset_categories}
This section defines the six categories used in ActMemEval. Each category reflects a different hidden constraint in the user's current request.
\begin{itemize}[leftmargin=*,itemsep=0pt,topsep=0pt,parsep=0pt]
\item \textbf{Safety-Health Risks:} the action may cause health, safety, toxic, material, or environmental risk.
\item \textbf{Feasibility Limits:} the action is impractical or cannot be carried out in reality.
\item \textbf{Time-Space-Procedure Mismatches:} the action conflicts with time, place, procedure, permission, policy, or order.
\item \textbf{Access or Availability Gaps:} a required resource, account, file, tool, item, or system is unavailable.
\item \textbf{Preference Incompatibility:} the plan conflicts with stable preferences, habits, soft constraints, or relationship context.
\item \textbf{Benefit-Reuse Opportunities:} an existing resource, right, reusable asset, or better option already exists.
\end{itemize}

Representative cases are shown below.

\categorycase{Safety-Health Risks}{It's cold in here. Search for tips on how to build a long-lasting wood fire in the fireplace.}{
\item Vet warned me that smoke can kill the cockatiel instantly.
\item Threw out all my non-stick pans and scented candles.
\item Checking the air quality monitor constantly for the bird's safety.
}{Urgent Stop; warn that smoke is fatal to the new parrot; suggest checking the thermostat instead.}

\categorycase{Feasibility Limits}{My nephew's birthday is coming up. Can you recommend some popular Lego sets suitable for a 12-year-old?}{
\item Nephew broke his arm and is in a cast.
\item He is bored because he cannot use his right hand well.
}{Legos might be a bit frustrating for him right now. Maybe a digital gift card, an audiobook, or a strategy board game would be better.}

\categorycase{Time-Space-Procedure Mismatches}{Schedule a team brainstorming workshop in Conference Room B for next Tuesday at 2 PM.}{
\item IT announced projector replacement next Tuesday afternoon.
\item The workshop requires a projector.
}{That time slot will not work for a presentation. Conference Room B is scheduled for projector replacement maintenance.}

\categorycase{Access or Availability Gaps}{I'm locked out of my Apple ID. Guide me on how to use the ``Recovery Key'' method to reset my password.}{
\item User generated a Recovery Key and printed it on paper.
\item User shredded all old loose papers during an office cleanup.
}{That might be a problem. Unless you saved a digital copy, you might have destroyed the key.}

\categorycase{Preference Incompatibility}{Book a table for two at ``The Butcher's Block'' steakhouse.}{
\item New partner is strict vegetarian.
\item Meat-only venues should be avoided.
}{Suggest an alternative restaurant with good vegetarian options for the new partner.}

\categorycase{Benefit-Reuse Opportunities}{I'm going to buy the ``Shadows of Rose'' DLC for Resident Evil Village on the PS5 store right now.}{
\item User bought the Gold Edition disc.
\item User redeemed the code inside the box.
}{You do not need to buy that. The Gold Edition already includes the DLC.}

\section{Human--LLM Agreement}
\label{sec:app_agreement}
For both retrieval accuracy and QA accuracy, we conduct a stratified sampling evaluation and randomly select 80 samples for each metric.
We then report Cohen's Kappa between two human annotators ($\kappa_{hh}$) and between the final human labels and the LLM judge ($\kappa_{hl}$), as shown in Table~\ref{tab:kappa_agreement}.
\begin{table}[t]
\centering
\small
\resizebox{0.8\linewidth}{!}{
\begin{tabular}{lcc}
\toprule
\textbf{Metric} & $\kappa_{hh}$ & $\kappa_{hl}$ \\
\midrule
Retrieval accuracy & 0.8956 & 0.8336 \\
QA accuracy & 0.8780 & 0.8410 \\
\bottomrule
\end{tabular}}
\caption{Cohen's Kappa scores for human--human and human--LLM agreement.}
\label{tab:kappa_agreement}
\end{table}
\section{Artifact License and Usage}
The ActMemEval dataset will be released under CC BY 4.0 and is intended for academic research use only.
\end{document}